\documentclass{article}
\pdfoutput=1
\usepackage{microtype}
\usepackage{graphicx}
\usepackage{booktabs} % for professional tables

\usepackage[utf8]{inputenc}
\usepackage[english]{babel}
\usepackage{amsmath,amsfonts,amssymb,amsthm, bm}

\usepackage{mathtools}
\usepackage{multicol,multirow}
\usepackage{caption}
\usepackage{subcaption}
\usepackage{wrapfig}

\usepackage{hyperref}

\usepackage{tikz}
\usetikzlibrary{fit,positioning,calc,matrix,shapes.multipart,backgrounds}

\usepackage{pgfplots}

\makeatletter
\renewcommand*\env@matrix[1][\arraystretch]{%
  \edef\arraystretch{#1}%
  \hskip -\arraycolsep
  \let\@ifnextchar\new@ifnextchar
  \array{*\c@MaxMatrixCols c}}
\makeatother

\newtheorem{lemma}{Lemma}
\newtheorem{remark}{Remark}

\DeclareMathOperator{\diag}{diag}
\newcommand{\SO}[1]{\mathrm{SO}(#1)}
\newcommand{\OO}[1]{\mathrm{O}(#1)}
\newcommand{\GL}{\mathrm{GL}}
\newcommand{\ZZ}{\mathbb{Z}}
\newcommand{\RR}{\mathbb{R}}
\newcommand{\II}{\mathbf{I}}
\newcommand{\CC}{\mathrm{C}}
\newcommand{\DD}{\mathrm{D}}
\newcommand{\KK}{\mathsf{K}}
\newcommand{\oKK}{\overline{\KK}}
\newcommand{\okappa}{\overline{\kappa}}
\newcommand{\KKC}{\KK^{\CC_N}}
\newcommand{\KKCzr}{\KK^{\CC_N}_{0 \rightarrow \text{reg}}}
\newcommand{\KKCkr}{\KK^{\CC_N}_{k \rightarrow \text{reg}}}
\newcommand{\KKChr}{\KKC_{\floor{\frac{N}{2}} \rightarrow \text{reg}}}
\newcommand{\KKCrr}{\KK^{\CC_N}_{\text{reg} \rightarrow \text{reg}}}
\newcommand{\KKD}{\KK^{\DD_N}}
\newcommand{\KKDzr}{\KK^{\DD_N}_{0 \rightarrow \text{reg}}}
\newcommand{\KKDzzr}{\KK^{\DD_N}_{0, 0 \rightarrow \text{reg}}}
\newcommand{\KKDzhr}{\KK^{\DD_N}_{0, \floor{\frac{N}{2}} \rightarrow \text{reg}}}
\newcommand{\KKDjkr}{\KK^{\DD_N}_{j, k \rightarrow \text{reg}}}

\newcommand{\KKDozr}{\KK^{\DD_N}_{1, 0 \rightarrow \text{reg}}}
\newcommand{\KKDohr}{\KK^{\DD_N}_{1, \floor{\frac{N}{2}} \rightarrow \text{reg}}}

\newcommand{\KKDrr}{\KK^{\DD_N}_{\text{reg} \rightarrow \text{reg}}}
\newcommand{\oKKC}{\oKK^{\CC_N}}
\newcommand{\oKKCkr}{\oKK^{\CC_N}_{k \rightarrow \text{reg}}}

\newcommand{\rhoDr}{\rho^{\DD_N}_\text{reg}}
\newcommand{\rhoCr}{\rho^{\CC_N}_\text{reg}}
\newcommand{\rhot}{\rho_\text{tri}}

\DeclarePairedDelimiter\floor{\lfloor}{\rfloor}

\usepackage[accepted]{icml2021}

\icmltitlerunning{FILTRA: Rethinking Steerable CNN by Filter Transform}

\begin{document}

\twocolumn[
\icmltitle{FILTRA: Rethinking Steerable CNN by Filter Transform}



\icmlsetsymbol{equal}{*}

\begin{icmlauthorlist}
\icmlauthor{Bo Li}{jd}
\icmlauthor{Qili Wang}{jd}
\icmlauthor{Gim Hee Lee}{nus}
\end{icmlauthorlist}

\icmlaffiliation{jd}{JD Technology}
\icmlaffiliation{nus}{National University of Singapore}

\icmlcorrespondingauthor{Bo Li}{prclibo@gmail.com}
% \icmlcorrespondingauthor{Eee Pppp}{ep@eden.co.uk}

\icmlkeywords{Machine Learning, ICML}

\vskip 0.3in
]



\printAffiliationsAndNotice{}  % leave blank if no need to mention equal contribution

\begin{abstract}
% Steerable CNN imposes the prior knowledge of transformation invariance or equivariance in the network architecture to enhance the the network robustness on geometry transformation of data and reduce overfitting. It has been an intuitive and widely used technique to construct a steerable filter by augmenting a filter with its transformed copies in the past decades, which is named as filter transform in this paper. Recently, the problem of steerable CNN has been studied from aspect of group representation theory, which reveals the function space structure of a steerable kernel function. However, it is not yet clear on how this theory is related to the filter transform technique. In this paper, we show that kernel constructed by filter transform can also be interpreted in the group representation theory. Meanwhile, we show that filter transformed kernels can be used to convolve input/output features in different group representation. This interpretation help complete the puzzle of steerable CNN theory and provides a novel and simple approach to implement steerable convolution operators. Experiments are executed on multiple datasets to verify the feasibility of the proposed approach.
Steerable CNN imposes the prior knowledge of transformation invariance or equivariance in the network architecture to enhance the the network robustness on geometry transformation of data and reduce overfitting. It has been an intuitive and widely used technique to construct a steerable filter by augmenting a filter with its transformed copies in the past decades, which is named as filter transform in this paper. Recently, the problem of steerable CNN has been studied from aspect of group representation theory, which reveals the function space structure of a steerable kernel function. However, it is not yet clear on how this theory is related to the filter transform technique. In this paper, we show that kernel constructed by filter transform can also be interpreted in the group representation theory. This interpretation help complete the puzzle of steerable CNN theory and provides a novel and simple approach to implement steerable convolution operators. Experiments are executed on multiple datasets to verify the feasibility of the proposed approach.

\end{abstract}

\section{Introduction}

Beyond the well-known property of equivariance under translation, there has been substantial recent interest in CNN architectures that are equivariant with respect to other transformation groups, e.g. reflection and rotation. Applications of such architectures range over scenarios where object orientation might variate, including OCR, aerial image processing, 3D point cloud processing, medical image processing, texture analysis and etc.

\subsection{Related Works}
Previous works on constructing equivariant CNN can be coarsely categorized as two aspects. The first aspect designs special \textit{steerable filters} so that the convolutional output is hard-baked to transform accordingly when the input reflects or rotates. A plenty of works develop this idea by filter rotation, including hand-crafted filters~\citep{oyallon2015deep} and learned filters~\citep{laptev2016ti, zhou2017oriented, cheng2018rotdcf, marcos2017rotation}. TI-Pooling~\citep{laptev2016ti} produce invariant output as input rotates. ORN~\citep{zhou2017oriented} and RotDCF~\citep{cheng2018rotdcf} produces output which circularly shifted as input rotates. Since each dimension of such permutable output corresponds to a uniformly discrete rotation angle, RotEqNet~\citep{marcos2017rotation} propose to extract rotation angle from the permutable features. 
%As will be described in Sect.~\ref{sec:main-results}, the invariant and permutable output features correspond to the concepts of \textit{trivial representation} and \textit{regular representation} in group representation theory, respectively. 
Another approach to construct steerable filters is to linearly combine a set of steerable bases. These bases can be solved in in the form of filter patches~\citep{cohen2014learning, cohen2016steerable}, harmonic bases~\citep{worrall2017harmonic, weiler2019general}, and differential operators~\cite{shen2020pdo}. \citet{weiler2019general}  comprehensively summarize works on steerable bases using polar Fourier basis.

The second aspect exploits specific transforms to act on input. Spatial Transformer Network (STN) is a  well-known representative, which predicts an affine matrix to transform its input to the canonical form. \citet{tai2019equivariant} inherits this idea to design equivariant network. 
Another choice of transform is to the polar coordinate system~\citep{henriques2017warped, esteves2018polar}. Since 2D rotation in Cartesian coordinate system corresponds to 2D translation in polar coordinate system, rotation equivariance can be achieved by conventional translation equivariant CNN.

\subsection{Contributions}
The approach proposed in this paper falls into the first category and is closely related to \citet{weiler2019general}. \citet{weiler2019general} exhaustively proves that all steerable convolutional operator could be denoted as the combination of a specific set of polar Fourier bases. However, it is not clear yet how this interpretation is related with the widely used filter transform technique. In this paper, we aim to establish the missing connection between the group representation based analysis for steerable filters and filter transform scheme. To this end, we propose a new approach (FILTRA) to use filter transform to establish steerability between features in different group representation in cyclic group $\CC_N$ and dihedral group $\DD_N$. The contribution of FILTRA can be summarized in two aspects:
\begin{itemize}
    \item \textbf{Theory.} FILTRA completes the missing link between the group representation based analysis for steerable filters and the filter transform technique. Since the latter technique is more widely known and used in the computer vision community, we believe this theory achievement will contribute to the application of steerable filters in computer vision research.
    \item \textbf{Practice.} We propose a novel approach to implement steerable convolutional operators which is equivalent to \citet{weiler2019general} but much simpler. For researchers who are not interested in the mathematical derivation, the boxed equations in this paper are the take-away tips which can be directly used to construct kernels in their CNN. As listed in the appendix, a minimal self-contained implementation only requires 60 lines of PyTorch code to realize all the functionality covered in this paper. The complete implementation will also be released as open source. Based on the proposed theory, we are able to analyze several conventional well-known filter transform approaches in a unified interpretation and show that FILTRA is a more general realization of the filter transform technique.
\end{itemize}

We verify the feasibility of FILTRA for the classification and regression tasks on different datasets. %We also show how the proposed steerable filters can be decomposed as polar Fourier bases. xxx

\section{Preliminaries}

% \subsection{Notations}

We make use of several NumPy or SciPy functions in equations including \texttt{roll}\footnote{\texttt{https://numpy.org/doc}}, \texttt{flipud}\footnote{\texttt{https://numpy.org/doc}} and \texttt{circulant}\footnote{\texttt{https://docs.scipy.org/doc/}}. We omit the variable in bracket sometimes by writing $\kappa^*_* = \kappa^*_*(g)$ and $\KK^*_* = \KK^*_*(\phi)$.

\begin{figure}
    \centering
    \input{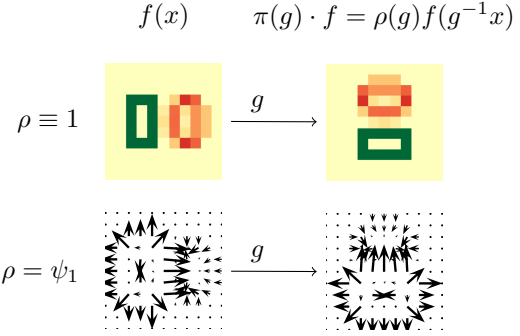}
    \caption{Examples of images (feature maps) with different group representation $\rho$. Both images undergo $90$deg rotation. The upper row is an RGB image whose 3-channel colors remain the same when the image is rotated. The lower row is a gradient image whose two channel value should be rotated in the same way when the gradient image is rotated.}
    \label{fig:vectorfield}
\end{figure}

\subsection{Steerable CNN}

We recapitulate the basic concepts of steerable CNN which will be frequently used in this paper. For detailed introduction, readers can refer to \citet{weiler2019general} for a comprehensive information. We mainly consider the 2D image case and denote $x \in \RR^2$ as a pixel coordinate. We use vector field $f(x) \in \RR^C$ to denote a general multi-channel image, where $C$ is the number of channels. Typical examples of $f(x)$ include RGB image $f(x) \in \RR^3$ and gradient image $f(x) \in \RR^2$. Consider a group $G$ of transformations and an element $g \in G$. Examples of $G$ include rotation, translation and flip. A vector field $f(x)$ follows the below rules when undergoing the act $\pi(g)$ of a group element $g$:
\begin{equation}
    \pi(g) \cdot f = \rho(g) f(g^{-1} x),
\end{equation}
where $\rho(g)$ is a group representation related to vector field $f$. Fig.~\ref{fig:vectorfield} shows an example of different types of $\rho$ for RGB images and gradient images under a rotation transform element $g$. The group representation of RGB is $\rho(g) \equiv 1$ while for gradient image $\rho(g)$ is a 2D rotation matrix which also rotates vector $f(x)$ by $g$.

In the scenario of convolutional neural network, a convolution operator $f \mapsto \kappa \cdot f$ is considered as \textit{steerable} if it satisfies
\begin{equation}
    \kappa \cdot [\pi_1(g) f] = \pi_2(g) [\kappa \cdot f],
\end{equation}
i.e. the output vector field transforms equivariantly under $g$ when the input is transformed by $g$.

\subsection{Reflection Group, Cyclic Group and Dihedral Group}

We consider steerable filters on reflection group $(\{\pm1\}, *)$, cyclic group $\CC_N$ and dihedral group $\DD_N = (\{\pm1\}, *) \ltimes \CC_N$. To unify the notations in derivation, we interpret $\CC_N = (\{1\}, *) \ltimes \CC_N$ and $(\{\pm1\}, *) = (\{\pm1\}, *) \ltimes \CC_1 = \DD_1$ so that a element in these three groups can always be denoted as a pair $g = (i_0, i_1)$, whose range is $\ZZ_2 \times \ZZ_1$ for reflection group, $\ZZ_1 \times \ZZ_N$ for cyclic group and $\ZZ_2 \times \ZZ_N$ for dihedral group. Each element in $\CC_N$ corresponds to rotation angle $\theta_{i_1} = \frac{2 i_1 \pi}{N}$.

\subsection{Group Representation}

A linear representation $\rho$ of a group $G$ on a vector space $\RR^n$ is a group homomorphism from G to the general linear group $\GL(n)$, denoted as
\begin{equation}\begin{gathered}
    \rho: G \mapsto \GL(n)\\
    \text{s.t.} \quad \rho(g g') = \rho(g) \rho(g'), \quad \forall g, g' \in G.
\end{gathered}\end{equation}
We consider three types of linear representation in this paper, i.e. trivial representation, regular representation and irreducible representation (irrep). Readers can refer to \citet{serre1977linear} for further background for these concepts.

The trivial representation of a group element is always $\rhot(g) \equiv 1$. The regular representation of a finite group $G$ acts on a vector space $\RR^{|G|}$ by permuting its axis. Therefore, for a rotation element $g = (0, i_1) \in \CC_N$ or $\DD_N$, we get
\begin{align}\begin{gathered}
    \label{eq:rot-perm}
    \rhoCr(g) = P(i_1), \quad
    \rhoDr(g) = \begin{bmatrix}
        P(i_1) & 0\\ 0 & P(i_1)\\
    \end{bmatrix}, \\\text{where} \quad
    P(i_1) = \texttt{roll}(\II_N, i_1, 0).
\end{gathered}\end{align}
For a reflected element $g = (1, i_1) \in \DD_N$, we get
\begin{align}\begin{gathered}
    \rhoDr(g) = \begin{bmatrix} 0 & B(i_1) \\ B(i_1) & 0 \end{bmatrix},
    \\\text{where} \quad
    B(i_1) = \texttt{flipud}(P(-i_1 - 1)).
\end{gathered}\end{align}

By selecting suitable change of basis of the vector space, a representation can be converted to a equivalent representation, which is the direct sum of several independent representations on the a series of orthogonal subspace. A representation is called irreducible representation if no non-trivial decomposition exists. This conversion is denoted as
\begin{equation}
    \rho(g) = Q \begin{bmatrix} \bigoplus_{(i_0, i_1) \in I} \psi_i(g) \end{bmatrix} Q^{-1},
\end{equation}
where $I$ is an index set specifying the irreducible representations $\psi_i$ and $Q$ is the change of basis.

\subsection{Filter Transform}

In this paper we use the term \textit{filter transform} to refer the technique to create a steerable filter by augmenting a conventional convolutional filter by its transformed copies. Generally the transform can be from reflection, cyclic or dihedral groups. Most previous works on filter transform studies the case of cyclic group. We denote the basic form of this rotated filter as $\KK$:
\begin{align}\begin{gathered}
    \KK(\phi) = \begin{bmatrix}\kappa^0 & \kappa^1 & \cdots & \kappa^{N - 1}\end{bmatrix}^\top,\\
    \kappa^n(\phi) = \kappa(\phi - \theta_n),\\
    \label{eq:filter-rotation}
\end{gathered}\end{align}
which is commonly used in previous works, e.g. TI-Pooling \citep{laptev2016ti}, ORN \citep{zhou2017oriented}, RotEqNet \citep{marcos2017rotation} and RotDCF~\citep{cheng2018rotdcf}. Based on the group representation theory, it is not difficult to note that the input to $\KK$ is trivial representation and the output is regular representation. If the input image is rotated by $\theta_1$, the concolved output features will circularly permute one step.

\subsection{Decomposing Regular Representation}
\label{sec:decomposing-regular}

\begin{figure}
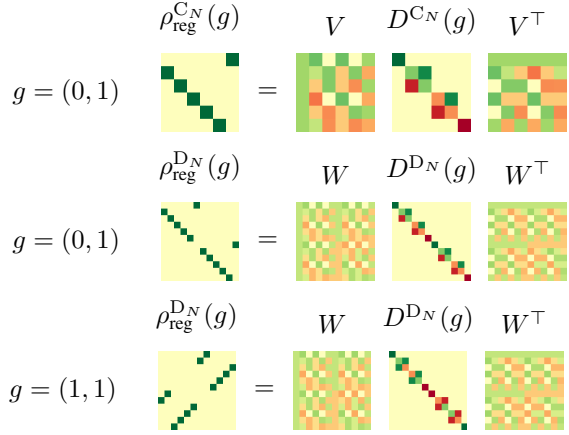

    \centering
    \input{figs/decomp_c6.tikz}\\
    \input{figs/decomp0.tikz}\\
    \input{figs/decomp1.tikz}
    \caption{Illustrations of \eqref{eq:cyclic-regular-decompose} under $\CC_6$ and \eqref{eq:dihedral-regular-decompose} under $\DD_6$. Red, light yellow and green denotes negative, 0 and positive matrix values, respectively. Note that each column in $V$ and $W$ corresponds to a discrete cosine transform basis.}
    \label{fig:decomp}
\end{figure}

We decompose the regular representation into a set of irreps. Define the following base irrep
\begin{align}\begin{gathered}
    \psi_{j, k}(i_0, i_1) = \Psi_k(i_1) \cdot F(i_0),\\
    \Psi_k(i_1) = \begin{cases}
        1 & \begin{matrix} k = 0 \text{ or } \\ k = \frac{N}{2}, \text{$N$ is even}\end{matrix}\\
        \begin{bmatrix}
            c[k\theta_{i_1}] & -s[k\theta_{i_1}]\\ s[k\theta_{i_1}] & c[k\theta_{i_1}]
        \end{bmatrix} & \text{otherwise}
    \end{cases},\\
    F(i_0) = \begin{cases}
        ((-1)^j)^{i_0}  & \begin{matrix} k = 0 \text{ or } \\ k = \frac{N}{2}, \text{$N$ is even}\end{matrix}\\
        \begin{bmatrix} 1 & 0 \\ 0 & (-1)^{i_0} \end{bmatrix} ((-1)^j)^{i_0} & \text{otherwise}
    \end{cases}.
\end{gathered}\end{align}
% \begin{align}\begin{split}
%     \psi_{j, k}(i_0, i_1) &=
%     \begin{cases}
%         ((-1)^j)^{i_0} & k = 0\\
%         (-1)^{i_1} \cdot ((-1)^j)^{i_0} & k = \frac{N}{2}, \text{$N$ is even}\\
%         \Psi \begin{bmatrix}
%         1 & 0 \\ 0 & (-1)^{i_0}
%         \end{bmatrix} \cdot ((-1)^j)^{i_0} &\text{otherwise}
%     \end{cases},\\
%     \Psi &= 
%         \begin{bmatrix}
%             c[k\theta_{i_1}] & -s[k\theta_{i_1}]\\ s[k\theta_{i_1}] & c[k\theta_{i_1}]
%         \end{bmatrix},
% \end{split}\end{align}
where $j, k$ are referred as the reflection and rotation frequency of the irrep. Concretely, if the action $g$ reflects/rotates an object once, $\psi_{j, k}(g)$ will reflects/rotates in vector space $j/k$ times. We also define the following discrete cosine transform basis
\begin{align}\begin{gathered}
    V = \begin{bmatrix}
    \beta_0 & \beta_1 & \cdots & \beta_{\floor{\frac{N}{2}}}
    \end{bmatrix},\\
    \beta_k = \begin{cases}
        \mathbf{1}_N & k = 0\\
        \begin{bmatrix}
        c[k \theta_0] \cdots & c[k \theta_{N - 1}]
        \end{bmatrix}^\top & k = \frac{N}{2}, \text{$N$ is even}\\
        \begin{bmatrix}[0]
        c[k \theta_0] \cdots & c[k \theta_{N - 1}]\\
        s[k \theta_0] \cdots & s[k \theta_{N - 1}]
        \end{bmatrix}^\top & \text{otherwise}
    \end{cases}.
\end{gathered}\end{align}
The following decomposition for $\rhoCr(0, i_1)$ holds
\begin{align}\begin{gathered}
    \rhoCr(g) = V D^{\CC_N} V^\top,\\
    D^{\CC_N} = \bigoplus_{0 \leq k \leq \floor{\frac{N}{2}}} \psi_{0, k}(0, i_1).
    \label{eq:cyclic-regular-decompose}
\end{gathered}\end{align}
The decomposition for $\rhoDr(i_0, i_1)$ holds in a bit more complicated form, i.e.
\begin{align}\begin{gathered}
    \rhoDr(i_0, i_1) = W D^{\DD_N} W^\top, \\
    W = \begin{bmatrix} V & V\\ V & -V \end{bmatrix},\\
    D^{\DD_N} = \bigoplus_{0 \leq j \leq 1, 0 \leq k \leq \floor{\frac{N}{2}}} \psi_{j, k}(i_0, i_1),
    \label{eq:dihedral-regular-decompose}
\end{gathered}\end{align}
and each column of $W$ is refered by $\beta_{j, k} = \begin{bmatrix}\beta_k^\top & (-1)^j \beta_k^\top\end{bmatrix}^\top$. See Fig.~\ref{fig:decomp} for a visualization of this decomposition.

\begin{remark}
    $V$ is actually a form of the discrete cosine transform matrix. \eqref{eq:cyclic-regular-decompose} and \eqref{eq:dihedral-regular-decompose} reveals that regular representation and irreducible representation is related by discrete cosine transform matrix. \eqref{eq:cyclic-regular-decompose} corresponds to the fact that a circulant matrix can be diagonalized by discrete Fourier transform matrix and \eqref{eq:dihedral-regular-decompose} can be verified by straight forward computation.
\end{remark}

We also mention a property of $\beta_k$ that is easy to verify and will be useful in our derivation.
\begin{subequations}
\begin{align}
\begin{split}
    \psi_{0, k}(0, i_1) \beta_k^\top &= \beta_k^\top P(i_1),\\
    \psi_{1, k}(0, i_1) \beta_k^\top &= \beta_k^\top P(i_1),\\% \quad g = (0, i_1)
\end{split}
\label{eq:beta-rotate0}\\
\begin{split}
    \psi_{0, k}(1, i_1) \beta_k^\top &= \beta_k^\top B(i_1),\\
    \psi_{1, k}(1, i_1) \beta_k^\top &= -\beta_k^\top B(i_1),\\% \quad g = (1, i_1).
\end{split}
\label{eq:beta-rotate1}
\end{align}
\end{subequations}
where $\psi_{0, k}(i_0, i_1)$ rotates column vectors of $\beta_k^\top$ as if they are circularly shifted.

% https://en.wikipedia.org/wiki/Circulant_matrix
% https://math.stackexchange.com/questions/297615/eigenvectors-of-a-circulant-matrix

\subsection{Harmonic Filters}

\citet{weiler20183d} proposes the condition of a filter kernel $\kappa$ to be equivariant under the action $g \in G$.
\begin{lemma}
The map $f \mapsto \kappa \cdot f$ is equivariant under G if and only if for all $g \in G$,
\begin{equation}
    \kappa(gx) = \rho_\text{out}(g) \kappa(x) {\rho_\text{in}(g)}^{-1}.
    \label{eq:kernel-constraints}
\end{equation}
\label{lm:condition}
\end{lemma}
\citet{weiler2019general} proves that such filters can be denoted by a series of harmonic bases $b(\phi)$, i.e. 
\begin{equation}
    \kappa(r, \phi) = \sum_{b \in \mathcal{K}} \omega_b(r) b(\phi),
    \label{eq:harmonic-denotion}
\end{equation}
where $\omega_b(r)$ is the per radial weights and $\mathcal{K}$ is a set of harmonic bases as dervied in the appendix of \citet{weiler2019general}. For example, consider $\rho_\text{in} = \psi_{i, m}$ and $\rho_\text{out} = \psi_{j, n}$ in $\DD_N$,
\begin{equation}\begin{split}
    \mathcal{K}_{\psi_{j, m} \leftarrow \psi_{i, n}} = \big\{ b_{\mu, \gamma, s}(\phi = \psi(\mu \phi) \xi(s) \big|\mu = m - s n, \\
    s \in \{\pm 1\} \big\}.
\end{split}\end{equation}

\section{Main Results}
\label{sec:main-results}
\eqref{eq:kernel-constraints} and \eqref{eq:harmonic-denotion} provide a general approach to verify and construct steerable CNN with different representations. In this section, we relate these theories with filter transform and show how to use filter transform to construct steerable filters with input/output of different representations. 

For readers who are not interested in group theory and mathematical derivation of the theory connection, we highlight the key equations to construct steerable filters in boxes. It should not be difficult to implement steerable filters directly from these equations using any modern deep learning framework. Fig.~\ref{fig:filters} shows illustration for these equations. 

In our derivation, we mainly consider the angular coordinate of polar coordinate functions $\kappa(r, \phi)$ and write them $\kappa(\phi)$. We will also frequently make use of the following property:
\begin{equation}
    \kappa(\phi - \theta_0) = \kappa(\phi + \theta_0), \quad
    \kappa(\phi - \theta_i) = \kappa(\phi + \theta_{N - i}).
    \label{eq:periodic}
\end{equation}

\subsection{From Trivial Representation to Regular Representation}

\paragraph{Rotation Group $\CC_N$}
Consider the the rotating filter $\KK$ in \eqref{eq:filter-rotation} and its reflected version $\oKK$:
\begin{align}\begin{gathered}
    \oKK(\phi) = \begin{bmatrix}\okappa^0 & \okappa^1 & \cdots & \okappa^{N - 1}\end{bmatrix}^\top,\\
    \okappa^n(\phi) = \kappa(\theta_n - \phi).
\end{gathered}\end{align}
The output of convolution with kernel $\KK$ or $\oKK$ naturally permutes as the input rotates in $\CC_N$. This intuitively corresponds to property of a steerable filter transforming from trivial representation to regular representation. In this paper, we use $\KK$ and $\oKK$ as the basic filters to construct different types of steerable filters in $\CC_N$ and $\DD_N$. We verify the observation of the above steerability by substituting $\KK$ into the LHS of Lemma~\ref{lm:condition} with $g = (0, 1)$ and write:
\begin{subequations}\begin{align}
    \label{eq:cn-trivial-proof}
    \KK(\phi + \theta_1) &= \begin{bmatrix}
        \kappa(\phi + \theta_1) & \kappa^0 & \cdots & \kappa^{N - 2}
        \end{bmatrix}^\top\\
        &= \begin{bmatrix}
        \kappa^{N - 1} & \kappa^0 & \cdots & \kappa^{N - 2}
        \end{bmatrix}^\top\\
        &= \rhoCr(0, 1) \KK {\rhot(0, 1)}^{-1}. \label{eq:cn-trivial-proof2}
\end{align}\end{subequations}
The above equation can be similarly verified for other $g = (0, i_1)$ and also on $\oKK$. Thus WLOG we select the steerable filter which transforms trivial representation to regular representation on $\CC_N$ as
\begin{equation}\boxed{
    \KKCzr = \KK.
}\label{eq:cn-trival-regular-kernel}
\end{equation}

\paragraph{Dihedral Group $\DD_N$}
The steerable filter that transforms trivial representation to regular representation on $\DD_N$ can be constructed as
\begin{equation}\boxed{
    \KKDzr(\phi) = \begin{bmatrix} \KK^\top & \oKK^\top \end{bmatrix}^\top,
}\label{eq:dn-trival-regular-kernel}
\end{equation}
which corresponds to enumerating each $\DD_N$ element and act on the kernel $\kappa$. For $g = (0, i_1)$, $\KKDzr$ can be verified to follow \eqref{eq:kernel-constraints} in the same way as \eqref{eq:cn-trivial-proof}, i.e. $\KKDzr(\phi + \theta) = \rhoDr(g) \KKDzr {\rhot(g)}^{-1}$.

For reflected action, when $g = (1, 1)$, we write:
\begin{align}\begin{split}
    \KK(-\phi + \theta_1) &= \left[\begin{matrix}
        \kappa(-\phi + \theta_1) & \kappa(-\phi - \theta_0)
    \end{matrix}\right.\\
    & \qquad \left.\begin{matrix}
        \kappa(-\phi - \theta_1) & \cdots & \kappa(-\phi - \theta_{N - 2})
    \end{matrix}\right]^\top\\
    &= \begin{bmatrix}
        \okappa^1 & \okappa^0 & \okappa^{N - 1} & \cdots & \okappa^2
    \end{bmatrix}^\top\\
    &= B(1) \oKK.
    \label{eq:trivial-exchange}
\end{split}\end{align}
Similarly, we can show for $g = (1, i_1)$,
\begin{align}\begin{gathered}
    \oKK(-\phi + \theta_{i_1}) = B(i_1) \KK,\\
    \KK(-\phi + \theta_{i_1}) = B(i_1) \oKK.
\end{gathered}\end{align}
Thus we verify \eqref{eq:kernel-constraints} for the reflected actions $g = (1, i_1)$ by summarizing the above as $\KKDzr(-\phi + \theta_{i_1}) = \rhoDr(g) \KKDzr {\rhot(g)}^{-1}$.

\begin{figure*}
    \centering
    \input{figs/filters.tikz}
    \caption{Visualization of FILTRA filter examples. Based on a same weight kernel $\KK$ (a Super Mario patch), we generate filters $\KKCzr$, $\KKDzr$, $\KKCkr$ and $\KKDjkr$. In this example we set $j=1, k=1, N=8$. The two-columns of matrix $\beta_k$ is splitted as $\beta_k^0$ and $\beta_k^1$ for visualization. Red, light yellow and green denotes negative, 0 and positive values, respectively. Please view this figure in color.}
    \label{fig:filters}
\end{figure*}

\subsection{From Irrep to Regular Representation}
\paragraph{Rotation Group $\CC_N$}
\label{sec:irrep-to-reg-rot}

Consider a $\CC_N$ irrep $\psi_{0,k}(g)$ with frequency $(0, k)$. We show that the following kernel
\begin{equation}\boxed{
    \KKCkr = \diag(\KK) \beta_k,
    \label{eq:irrep-cn-kernel}
}\end{equation}
transforms from $\psi_{0,k}(g)$ to regular representation for the action $g = (0, i_1)$. The derivation of correctness can be found in the appendix.

\paragraph{Dihedral Group $\DD_N$}

Consider a $\DD_N$ irrep $\psi_{j,k}(i_0, i_1)$ with frequency $(j, k)$. We show that the following kernel:
\begin{equation}\boxed{
    \KKDjkr = %\diag(\KKD) \beta_{j, k} =
    % \begin{cases}
    %     \begin{bmatrix}{\KKCkr}^\top & {\oKK^{\CC_N}_{k \rightarrow \text{reg}}}^\top\end{bmatrix}^\top & j = 0\\
    %     \begin{bmatrix}{\KKCkr}^\top & -{\oKK^{\CC_N}_{k \rightarrow \text{reg}}}^\top\end{bmatrix}^\top & j = 1
    % \end{cases}.
    \begin{bmatrix}{\KKCkr}^\top & (-1)^j \cdot {\oKKCkr}^\top\end{bmatrix}^\top
}\label{eq:irrep-dn-kernel}
\end{equation}
transforms from $\psi_{j,k}(i_0, i_1)$ to regular representation for the action $g = (i_0, i_1) \in \DD_N$. 

\subsection{From Regular Representation to Regular Representation}
\label{sec:reg-reg-cn}

Regular representation possesses a nice property that it can be averaged, pooled or activated channel-wise without violating steerability \citep{weiler2019general}. Thus it is convenient to used regular representation for the intermediate features of a steerable CNN. We show in this subsection that the following kernels can be use to construct a steerable kernel whose input and output features are both in regular representation.

\paragraph{Rotation Group $\CC_N$}
\begin{equation}\boxed{
    \KKCrr = \begin{bmatrix} \KKCzr \cdots \KKChr \end{bmatrix} V^{-1}.
}\label{eq:cn-regular-regular-kernel}
\end{equation}

\paragraph{Dihedral Group $\DD_N$}

\begin{equation}\boxed{
        \begin{split}
            \KKDrr &= \left[\begin{matrix} \KKDzzr \cdots \KKDzhr \end{matrix}\right.\\
                &\qquad\left.\begin{matrix}\KKDozr \cdots \KKDohr  \end{matrix}\right] W^{-1}.
        \end{split}
}\label{eq:dn-regular-regular-kernel}
\end{equation}

The above two kernels can be verified to transform regular representation to regular representation in similar way and we show the derivation for the $\CC_N$ case \eqref{eq:cn-regular-regular-kernel} as an example in the appendix.

\subsection{Reversed Transform of Representations}

It is obvious to find that for \eqref{eq:kernel-constraints}, if $\rho_\text{in}, \rho_\text{out}$ are orthogonal matrices, i.e. $\rho_\text{in}^{-1} = \rho_\text{in}^\top, \rho_\text{out}^{-1} = \rho_\text{out}^\top$, the transpose of \eqref{eq:kernel-constraints} naturally proves the equivariance of $\kappa^\top$ under a reversed representation transform direction, i.e. from $\rho_\text{out}$ to $\rho_\text{in}$. Thus we can easily obtain equivariance kernel from regular representation to trivial/irreducible representation by simply transposing \eqref{eq:cn-trival-regular-kernel}, \eqref{eq:dn-trival-regular-kernel}, \eqref{eq:irrep-cn-kernel} and \eqref{eq:irrep-dn-kernel}.

\subsection{Conventional Rotating Filters}

We comprehensively study the approach to use filter rotation to form steerable convolutional kernels with regular representation features as input or output. Conventional filter rotation based networks exploit some basic forms introduced in this section. TI-Pooling~\citep{laptev2016ti} exploits kernel $\KKC$ to transform trivial to regular representation, executes orientation pooling to convert regular to trivial representation and loses orientation information. RotDCF and ORN exploits a kernel of form
\begin{equation}
    \KKC_\text{ORN} = \texttt{circulant}(\KK).
\end{equation}
It is easy to verify that $\KKC_\text{ORN}$ also follows Lemma~\ref{lm:condition} to be a steerable filter. However, compared to $\KKCrr$, $\KKC_\text{ORN}$ consumes same filter storage but has less weight capacity ($N$ v.s. $N \floor{\frac{N}{2}}$). RotEqNet constructs 2D vector field which could rotate as its input rotates but regards the 2D vector field as independent trivial representation in convolution. As shown in this paper, it preserves better steerability to regards the vector field as irrep representation with frequency $1$.

% \subsection{Relationship with $\SO2$ and $\OO2$}

\subsection{Numerical Accuracy for Discrete Kernels}
\label{sec:numerical}
Note that when implementing discrete convolution, the equality of \eqref{eq:cn-trivial-proof} does not perfectly hold. For example, consider $\kappa^n(\phi) = \kappa(\phi - \theta_n)$, $\kappa^n(\theta_n)=\kappa(0)$ holds for a continuous $\kappa$. However, for discrete $\kappa$, $\kappa^n(\phi)$ is a rotated interpolation of $\kappa(\phi)$ and this equality does not precisely hold in general. However, in our implementation this is not specifically handled since the overall performance does not drop too much.

There exist some exceptions where the equality can be achieved for discrete $\kappa$. One example is when $\kappa^n(\phi)$ is a $90^\circ$ rotation of $\kappa$ and it can be precisely constructed from $\kappa$. Another example is when $\kappa^n$ is a $45^\circ$ rotation interpolated by nearest pixel from a $\kappa$ of size $3\times3$.

\subsection{Relation with Continuous Rotation}

Consider \eqref{eq:cyclic-regular-decompose}, obviously $\rhoCr(0, i_1)$ is a continuous function of $i_1$. If $i_1$ is a real number in $(i, i + 1)$, we can observe that $\rhoCr(0, i_1)$ will smoothly interpolates between $\rhoCr(0, i)$ and $\rhoCr(0, i + 1)$. This interpolation is not linear. However, in practice we found that for $N \leq 4$ this interpolation approximately captures the continuity of rotation well in the outputted regular represented features.

\subsection{Steerable CNN with Multiple Layers}

A conventional CNN is usually composed convolution, pooling, nonlinearity and fully-connected layers. To achieve equivariance for the overall network, it is desired that all the component layers are steerable. As analyzed in the appendix of \citet{weiler2019general}, channel-wise nonlinearity and channel-wise pooling preserves the steerability on feature maps with regular representation. fully-connected layers is a special case of convolution with $1 \times 1$ kernels and thus can be easily realized by steerable convolution.

\section{Experiments}

\begin{table*}[]
    \footnotesize
    \caption{Network structure in experiments}
    \centering
    \begin{subtable}[t]{0.55\linewidth}
        \centering
        \vspace{0pt}
        \caption{The backbone network structure used in our experiments is composed by convolution, ReLU and pooling layers. The convolution layers are realized by FILTRA, R2Conv and conventional convolution respectively while the rest layers remain the same. Three realizations have the same number of output channels in each layer but organize the channels to be follow regular representation for FILTRA and R2Conv. k: kernel size. s: stride. $\delta t$: filter generation time in ms.}
        \begin{tabular}{lrrrrr}
            \hline
            layer & k & s & output & $\delta t$ (FIL) & $\delta t$ (R2)\\
            \hline
            conv+relu & $5$ & $1$ & $128$ (reg) & 0.12 & 0.17\\
            conv+relu & $5$ & $1$ & $192$ (reg) & 0.13 & 0.13\\
            pool & $3$ & $2$ & $256$ (reg) & - & -\\
            conv+relu & $5$ & $1$ & $256$ (reg) & 0.13 & 0.13\\
            conv+relu & $5$ & $1$ & $384$ (reg) & 0.23 & 0.23\\
            pool & $3$ & $2$ & $384$ (reg) & - & -\\
            conv+relu & $5$ & $1$ & $512$ (reg) & 0.32 & 0.48\\
            conv+relu & $5$ & $1$ & $768$ (reg) & 0.62 & 0.91\\
            pool & $3$ & $2$ & $768$ (reg) & - & -\\
            \hline
        \end{tabular}
        \label{tab:backbone}
    \end{subtable}\quad
    \vspace{0pt}
    \begin{subtable}[t]{0.40\linewidth}
        \centering
        \vspace{0pt}
        \caption{The classification head network structure used in our experiments uses a Grouping Pooling~\citep{weiler2019general} to generate transform invariant features.}
        \begin{tabular}{lrrr}
            \hline
            layer & k & s & output\\
            \hline
            GroupPool & $-$ & $1$ & $24$ (reg) \\
            fc+relu & $-$ & $-$ & $16$ (reg) \\
            fc & $-$ & $-$ & $10$ (tri) \\
            \hline
        \end{tabular}
        \label{tab:classification-head}
    % \end{subtable}
    % \begin{subtable}{0.40\linewidth}
        \caption{The regression head network structure used in our experiments uses a PointwiseAdaptiveMaxPool~(PAMaxPool)~\citep{weiler2019general} to summarize feature in regular representation.% We use a $1 \times 1$ kernel convolution as fully-connected layers and realize them as the three experimented convolutional operators respectively.
        }
        \begin{tabular}{lrrrrr}
            \hline
            layer & k & s & output \\
            \hline
            PAMaxPool & $-$ & $-$ & $24$ (reg) \\
            conv+relu & $1$ & $1$ & $16$ (reg)\\
            conv & $1$ & $1$ & $2$ (irrep)\\
            \hline
        \end{tabular}
        \label{tab:regression-head}
    \end{subtable}
    \label{tab:network}
\end{table*}

% \subsection{Setup}

The proposed equivariant convolution, refered as FILTRA, can be interpreted as an alternative formulation for the harmonic based \citep{weiler2019general} implementation of steerable convolution. In this section we show the pros and cons of each implementation by experiments. We make use of the framework E2CNN~\citep{weiler2019general} for our experiments as it provides the general interface and operations for steerable CNN network. Experiments are executed on the MNIST, KMNIST~\citep{clanuwat2018deep}, FashionMNIST~\citep{xiao2017fashion}, EMNIST~\citep{cohen2017emnist} and CIFAR10 datasets. 

We compare FILTRA against two convolution operations, i.e. the representative harmonic based convolution R2Conv~\citep{weiler2019general} from E2CNN and the conventional vanilla convolution. All MNIST-like datasets are experimented on a same feature extraction backbone as described in Table~\ref{tab:backbone}, with convolution operator realized by the three experimented approaches. CIFAR10 is experimented with WideResNet \citep{zagoruyko2016wide} in the setting similar to \cite{weiler2019general}. $\CC_8$ and $\DD_8$ steerable networks are used in the experiments. For all experiments, we randomly rotate or reflect according to the experiment settings. The settings and evaluation results are listed in Table~\ref{tab:eval}. Different from~\citet{weiler20183d}, we force the three convolution kernels to output same number of channels. For example, compared to vanilla convolution, the number of free weights for a $\CC_8$ FILTRA is reduced to $1/8$ and for a $\DD_8$ is reduced to $1/16$. The filters for all the approaches will thus have exactly same shape at the deploy stage.

Experiments are executed on GTX 2070. The training procedure of FILTRA and R2Conv can both be implemented as a vanilla convolution plus a filter generation step. For $\CC_8$ case the runtime of both approaches is similar and for $\DD_8$ case FILTRA is slightly faster. We show runtime of $\DD_8$ case in Table~\ref{tab:backbone} at training stage. R2Conv additionally requires a initialization of about $2$ min. Both of the approaches consume the same inference time as of vanilla convolution.

Our implementation of FILTRA is released on Github\footnote{\url{github.com/prclibo/filtra}}.

\subsection{Classification Task}

The most typical experiment used in previous works on conventional steerable CNN is the classification task. We follow this convention and compare the classification performance of the experimented three approaches in Table~\ref{tab:eval}. FILTRA show comparable performance to R2Conv and slightly improves accuracy for OCR-like (*MNIST) tasks where high frequency texture is limited. On CIFAR10, the performance of FILTRA is minorly disadvantageous. The explanation comes in the interpolation artifacts mentioned in Subsect.~\ref{sec:numerical}. As the interpolation of high frequency components deviates more, this harms the performance on CIFAR10 with high frequency texture.

\begin{table*}[]
    \footnotesize
    \caption{Performance on MNIST and CIFAR10. S: randomly augmented over $\SO2$. O: randomly augmented over $\OO2$. wrn: WideResNet. \citet{zagoruyko2016wide}.}
    \setlength{\tabcolsep}{2pt}
    \centering
    \begin{tabular}{l|rr|rr|rr|rr|rr|rr|rr|rr|rr}
        \hline
        \multirow{2}{*}{Tasks} & \multicolumn{10}{c}{Classification (acc)} & \multicolumn{8}{c}{Regression (angle err deg)} \\
        \cline{2-19}
        & \multicolumn{2}{c}{mnist}  & \multicolumn{2}{c}{kmnist} & \multicolumn{2}{c}{fmnist} & \multicolumn{2}{c}{emnist} & \multicolumn{2}{c}{cifar10} & \multicolumn{2}{c}{mnist} & \multicolumn{2}{c}{kmnist} & \multicolumn{2}{c}{fmnist} & \multicolumn{2}{c}{emnist}\\
        % \hline
        % Backbone & b5 & b5 & b5 & b5 & b5 & b5 & b5 & b5 & wrn & wrn & b5 & b5 & b5 & b5 & b5 & b5 & b5 & b5 \\
        \hline
        Aug& S & O & S & O & S & O & S & O & wrn & wrn & S & O & S & O & S & O & S & O \\
        \hline
        Net eqiv & $\CC_8$ & $\DD_8$ & $\CC_8$ & $\DD_8$ & $\CC_8$ & $\DD_8$ & $\CC_8$ & $\DD_8$ & $\CC_8$ & $\DD_8$ & $\CC_8$ & $\DD_8$ & $\CC_8$ & $\DD_8$ & $\CC_8$ & $\DD_8$ & $\CC_8$ & $\DD_8$ \\
        \hline
        FILTRA & 98.9 & 98.1 & 97.1 & 97.0 & 90.5 & 90.8 & 77.1 & 80.5 & 93.4 & 92.8 & 3.3 & 5.4 & 3.2 & 3.6 & 2.6 & 2.8 & 29.8 & 24.9\\
        R2Conv & 98.8 & 98.1 & 97.3 & 96.8 & 90.5 & 90.8 & 76.7 & 80.1 & 93.6 & 92.7 & 4.8 & 8.9 & 3.4 & 4.5 & 2.9 & 3.7 & 34.5 & 29.2\\
        Conv & 98.5 & 98.0 & 96.4 & 95.2 & 89.3 & 88.3 & 72.6 & 80.1 & 93.2 & - & 6.6 & 10.6 & 4.8 & 6.4 & 3.1 & 3.6 & 37.4 & 25.5\\
        \hline
    \end{tabular}
    \label{tab:eval}
\end{table*}

% \begin{table}[]
%     \setlength{\tabcolsep}{3pt}
%     \centering
%     \begin{tabular}{l|rrr|rrr|rrr|rrr|rrr}
%         \hline
%         \multirow{2}{*}{Tasks} & \multicolumn{9}{c}{Classification (acc)} & \multicolumn{6}{c}{Regression (angle err deg)} \\
%         \cline{2-16}
%         & \multicolumn{3}{c}{MNIST}  & \multicolumn{3}{c}{KMNIST} & \multicolumn{3}{c}{CIFAR10} & \multicolumn{3}{c}{MNIST} & \multicolumn{3}{c}{KMNIST} \\
%         \hline
%          Setting & ccc & dcc & ddd & ccc & dcc & ddd & cfi & dfi & ddi & ccc & dcc & ddd & ccc & dcc & ddd \\
%         \hline
%         FILTRA & 98.9 & 98.1 & 98.1 & 97.1 & 96.9 & 97.0 & 99.9 & 99.9 & 99.9 & 3.3 & 5.4 & 5.4 & 3.2 & 4.1 & 3.6\\
%         R2Conv & 98.8 & 98.1 & 98.1 & 97.3 & 96.7 & 96.8 & 99.9 & 99.9 & 99.9 & 4.8 & 8.6 & 8.9 & 3.4 & 4.6 & 4.5\\
%         Conv & 98.5 & - & 98.0 & 96.4 & - & 95.2 & 99.9 & 99.9 & 99.9 & 6.6 & - & 10.6 & 4.8 & - & 6.4\\
%         \hline
%     \end{tabular}
%     \caption{Performance on MNIST and CIFAR10. Setting letter triplets indicate the network equivariance, training data augmentation and testing data augmentation (c: $\CC_8$, d: $\DD_8$, f: $\DD_1$, i: identity). }
%     \label{tab:eval}
% \end{table}

\subsection{Regression Task}

Besides the typical classification task, we find that the property of steerability is naturally advantageous for many regression tasks whose input might rotate or reflect. In this paper, we evaluate the regression performance with an example task to predict the character direction. Similar tasks are commonly used in OCR techniques. When the character rotates, the predicted direction should rotate with the same rotating frequency. This means the predicted 2D direction vector is following a irrep $\psi_{0, 1}$ for $\CC_N$. We reuse the backbone in Table~\ref{tab:backbone} to extract features and use a regression head in Table~\ref{tab:regression-head} to predict a unit 2D vector denoting the direction. The network is trained with MSE loss. Note that the images should be masked by a disk to avoid the network to overfit the direction from rotated black boundary. Different approaches are evaluated by the mean included angle between the predicted and groundtruth directions as shown in Table~\ref{tab:eval}. FILTRA with $\CC_8$ steerability performs best when trained on data augmented over $\SO2$. We owe this to the fact that FILTRA weight is naturally organized by the discrete grid layout. Each element of discrete weight matrix contribute to one more DoF of the filters. In contrast, R2Conv uses filters parameterized with a polar coordinate. The DoF of the filters is slightly reduced due to the discretization.

\section{Conclusions}
In this paper, we establish the connection between the recent steerable CNN structure based on group representation theory and the conventional transformed filters. To this end, we propose an approach to construct steerable convolution filters, which transform between features in trivial, irreducible and regular representations. We verify the feasibility of FILTRA for the classification and regression tasks on several datasets.


\bibliography{main}

\begin{thebibliography}{19}
\providecommand{\natexlab}[1]{#1}
\providecommand{\url}[1]{\texttt{#1}}
\expandafter\ifx\csname urlstyle\endcsname\relax
  \providecommand{\doi}[1]{doi: #1}\else
  \providecommand{\doi}{doi: \begingroup \urlstyle{rm}\Url}\fi

\bibitem[Cheng et~al.(2018)Cheng, Qiu, Calderbank, and Sapiro]{cheng2018rotdcf}
Cheng, X., Qiu, Q., Calderbank, R., and Sapiro, G.
\newblock Rotdcf: Decomposition of convolutional filters for
  rotation-equivariant deep networks.
\newblock \emph{arXiv preprint arXiv:1805.06846}, 2018.

\bibitem[Clanuwat et~al.(2018)Clanuwat, Bober-Irizar, Kitamoto, Lamb, Yamamoto,
  and Ha]{clanuwat2018deep}
Clanuwat, T., Bober-Irizar, M., Kitamoto, A., Lamb, A., Yamamoto, K., and Ha,
  D.
\newblock Deep learning for classical japanese literature.
\newblock \emph{arXiv preprint arXiv:1812.01718}, 2018.

\bibitem[Cohen et~al.(2017)Cohen, Afshar, Tapson, and
  Van~Schaik]{cohen2017emnist}
Cohen, G., Afshar, S., Tapson, J., and Van~Schaik, A.
\newblock Emnist: Extending mnist to handwritten letters.
\newblock In \emph{2017 International Joint Conference on Neural Networks
  (IJCNN)}, pp.\  2921--2926. IEEE, 2017.

\bibitem[Cohen \& Welling(2014)Cohen and Welling]{cohen2014learning}
Cohen, T. and Welling, M.
\newblock Learning the irreducible representations of commutative lie groups.
\newblock In \emph{International Conference on Machine Learning}, pp.\
  1755--1763, 2014.

\bibitem[Cohen \& Welling(2016)Cohen and Welling]{cohen2016steerable}
Cohen, T.~S. and Welling, M.
\newblock Steerable cnns.
\newblock \emph{arXiv preprint arXiv:1612.08498}, 2016.

\bibitem[Esteves et~al.(2018)Esteves, Allen-Blanchette, Zhou, and
  Daniilidis]{esteves2018polar}
Esteves, C., Allen-Blanchette, C., Zhou, X., and Daniilidis, K.
\newblock Polar transformer networks.
\newblock In \emph{International Conference on Learning Representations}, 2018.

\bibitem[Henriques \& Vedaldi(2017)Henriques and Vedaldi]{henriques2017warped}
Henriques, J.~F. and Vedaldi, A.
\newblock Warped convolutions: Efficient invariance to spatial transformations.
\newblock In \emph{International Conference on Machine Learning}, pp.\
  1461--1469. PMLR, 2017.

\bibitem[Laptev et~al.(2016)Laptev, Savinov, Buhmann, and
  Pollefeys]{laptev2016ti}
Laptev, D., Savinov, N., Buhmann, J.~M., and Pollefeys, M.
\newblock Ti-pooling: transformation-invariant pooling for feature learning in
  convolutional neural networks.
\newblock In \emph{Proceedings of the IEEE conference on computer vision and
  pattern recognition}, pp.\  289--297, 2016.

\bibitem[Marcos et~al.(2017)Marcos, Volpi, Komodakis, and
  Tuia]{marcos2017rotation}
Marcos, D., Volpi, M., Komodakis, N., and Tuia, D.
\newblock Rotation equivariant vector field networks.
\newblock In \emph{Proceedings of the IEEE International Conference on Computer
  Vision}, pp.\  5048--5057, 2017.

\bibitem[Oyallon \& Mallat(2015)Oyallon and Mallat]{oyallon2015deep}
Oyallon, E. and Mallat, S.
\newblock Deep roto-translation scattering for object classification.
\newblock In \emph{Proceedings of the IEEE Conference on Computer Vision and
  Pattern Recognition}, pp.\  2865--2873, 2015.

\bibitem[Serre(1977)]{serre1977linear}
Serre, J.-P.
\newblock \emph{Linear representations of finite groups}, volume~42.
\newblock Springer, 1977.

\bibitem[Shen et~al.(2020)Shen, He, Lin, and Ma]{shen2020pdo}
Shen, Z., He, L., Lin, Z., and Ma, J.
\newblock Pdo-econvs: Partial differential operator based equivariant
  convolutions.
\newblock In \emph{International Conference on Machine Learning}, pp.\
  8697--8706. PMLR, 2020.

\bibitem[Tai et~al.(2019)Tai, Bailis, and Valiant]{tai2019equivariant}
Tai, K.~S., Bailis, P., and Valiant, G.
\newblock Equivariant transformer networks.
\newblock In \emph{International Conference on Machine Learning (ICML)}, 2019.

\bibitem[Weiler \& Cesa(2019)Weiler and Cesa]{weiler2019general}
Weiler, M. and Cesa, G.
\newblock General e (2)-equivariant steerable cnns.
\newblock In \emph{Advances in Neural Information Processing Systems}, pp.\
  14334--14345, 2019.

\bibitem[Weiler et~al.(2018)Weiler, Geiger, Welling, Boomsma, and
  Cohen]{weiler20183d}
Weiler, M., Geiger, M., Welling, M., Boomsma, W., and Cohen, T.~S.
\newblock 3d steerable cnns: Learning rotationally equivariant features in
  volumetric data.
\newblock In \emph{Advances in Neural Information Processing Systems}, pp.\
  10381--10392, 2018.

\bibitem[Worrall et~al.(2017)Worrall, Garbin, Turmukhambetov, and
  Brostow]{worrall2017harmonic}
Worrall, D.~E., Garbin, S.~J., Turmukhambetov, D., and Brostow, G.~J.
\newblock Harmonic networks: Deep translation and rotation equivariance.
\newblock In \emph{Proceedings of the IEEE Conference on Computer Vision and
  Pattern Recognition}, pp.\  5028--5037, 2017.

\bibitem[Xiao et~al.(2017)Xiao, Rasul, and Vollgraf]{xiao2017fashion}
Xiao, H., Rasul, K., and Vollgraf, R.
\newblock Fashion-mnist: a novel image dataset for benchmarking machine
  learning algorithms.
\newblock \emph{arXiv preprint arXiv:1708.07747}, 2017.

\bibitem[Zagoruyko \& Komodakis(2016)Zagoruyko and
  Komodakis]{zagoruyko2016wide}
Zagoruyko, S. and Komodakis, N.
\newblock Wide residual networks.
\newblock \emph{arXiv preprint arXiv:1605.07146}, 2016.

\bibitem[Zhou et~al.(2017)Zhou, Ye, Qiu, and Jiao]{zhou2017oriented}
Zhou, Y., Ye, Q., Qiu, Q., and Jiao, J.
\newblock Oriented response networks.
\newblock In \emph{Proceedings of the IEEE Conference on Computer Vision and
  Pattern Recognition}, pp.\  519--528, 2017.

\end{thebibliography}
\bibliographystyle{icml2021}


\end{document}

\maketitle

\section{Verification of Lemma~\ref{lm:condition} on \eqref{eq:irrep-cn-kernel}}
\eqref{eq:irrep-cn-kernel} can be verified to follow Lemma~\ref{lm:condition} as:
\begin{align}\begin{split}
    \label{eq:cn-irrep-proof}
    &\KKCkr(\phi + \theta_{i_1}) = \diag\big(P(i_1) \KK\big) \beta_k, \quad \text{c.f. \eqref{eq:cn-trivial-proof2}}\\
    &= P(i_1) \diag(\KK) P(i_1)^{-1} \beta_k = \rhoCr(g) \KKCkr \psi_{0, k}(g)^{-1}, \quad \text{c.f. \eqref{eq:beta-rotate0}.}
\end{split}\end{align}
We can also verify this for
\begin{equation}
    \oKKCkr = \diag(\oKK) \beta_k.
    \label{eq:irrep-cn-kernel-conj}
\end{equation}

\section{Verification of Lemma~\ref{lm:condition} on \eqref{eq:irrep-dn-kernel}}

First note it is easy to verify that for $i_0 = 0$, i.e. $g = (0, i_1)$, the Lemma~\ref{lm:condition} holds in the same way as \eqref{eq:cn-irrep-proof},
\begin{align}
    \KKDjkr(\phi + \theta) &= \rhoDr(g) \KKDjkr {\psi_{j, k}(g)}^{-1}.
\end{align}
We then generalize \eqref{eq:trivial-exchange} on $\KKCkr$ and $\oKK^{\CC_N}_{k \rightarrow \text{reg}}$ given a reflected action $g = (1, i_1)$:

\begin{subequations}
\begin{align}
    \KKCkr(-\phi + \theta_{i_1}) &= \diag\big(\KK(-\phi + \theta_{i_1})\big) \beta_k = B(i_1) \diag(\oKK) B(i_1)^{-1} \beta_k, \quad \text{c.f. \eqref{eq:beta-rotate1}}\\
    &= B(i_1) \diag(\oKK) \beta_k \psi_{0, k}(g)^{-1} = -B(i_1) \diag(\oKKC) \beta_k \psi_{1, k}(g)^{-1}
    \label{eq:irrep-exchange}\\
    %&\oKKCkr(-\phi + \theta_{i_1}) 
    &= B(i_1) \diag(\KKC) \beta_k \psi_{0, k}(g)^{-1} = -B(i_1) \diag(\KKC) \beta_k \psi_{1, k}(g)^{-1}.
    \label{eq:irrep-exchange-conj}
\end{align}\end{subequations}

Note that \eqref{eq:irrep-exchange} and \eqref{eq:irrep-exchange-conj} both have two equivalent forms denoted with $\psi_{0, k}(g)$ and $\psi_{1, k}(g)$ respectively. Now we can show $\KK^{\DD_N}_{j, k \rightarrow \text{reg}}$ follows Lemma~\ref{lm:condition} for $j=0$, $i_0=1$, i.e. $g = (1, i_1)$ as:
\begin{subequations}
\begin{align}
    &\KKDjkr(-\phi + \theta_{i_1}) =\begin{bmatrix}{\KKCkr(-\phi + \theta_{i_1})}^\top & {\oKKCkr}(-\phi + \theta_{i_1})^\top\end{bmatrix}^\top\\
    &= \begin{bmatrix}B(i_1) \diag(\oKKC) \beta_k \psi_{0, k}(g)^{-1} & B(i_1) \diag(\KKC) \beta_k \psi_{0, k}(g)^{-1}\end{bmatrix}^\top \quad\text{c.f. \eqref{eq:irrep-exchange}}\\
    &= \rhoDr(g) \begin{bmatrix}{\KKCkr}^\top & {\oKKCkr}^\top\end{bmatrix}^\top \psi_{0, k}(g)^{-1} \quad\text{c.f. \eqref{eq:irrep-cn-kernel}, \eqref{eq:irrep-cn-kernel-conj}}\\
    &= \rhoDr(g) \KKDjkr \psi_{0, k}(g)^{-1}.
\end{align}
\end{subequations}
The verification is similar for $j=1$, $i_0=1$, i.e. $g = (1, i_1)$:
\begin{subequations}
\begin{align}
    &\KKDjkr(-\phi + \theta_{i_1}) =\begin{bmatrix}{\KKCkr(-\phi + \theta_{i_1})}^\top & -{\oKKCkr}(-\phi + \theta_{i_1})^\top\end{bmatrix}^\top\\
    &= \begin{bmatrix} -B(i_1) \diag(\oKKC) \beta_k \psi_{1, k}(g)^{-1} & B(i_1) \diag(\KKC) \beta_k \psi_{1, k}(g)^{-1}\end{bmatrix}^\top \quad\text{c.f. \eqref{eq:irrep-exchange}}\\
    &= \rhoDr(g) \begin{bmatrix}{\KKCkr}^\top & -{\oKKCkr}^\top\end{bmatrix}^\top \psi_{0, k}(g)^{-1} \quad\text{c.f. \eqref{eq:irrep-cn-kernel}, \eqref{eq:irrep-cn-kernel-conj}}\\
    &= \rhoDr(g) \KKDjkr \psi_{0, k}(g)^{-1}.
\end{align}
\end{subequations}

\section{Verification of Lemma~\ref{lm:condition} on \eqref{eq:cn-regular-regular-kernel}}

This kernel can be verified as follows for $g = (0, i_1)$:
\begin{subequations}
\begin{align}
    &\KKCrr(\phi + \theta_{i_1}) = \begin{bmatrix} \rhoCr(g) \KKCzr \psi_{0, 0}(g)^{-1}, \cdots, \rhoCr(g) \KKChr \psi_{0, \floor{\frac{N}{2}}}(g)^{-1} \end{bmatrix} V^{-1}\\
    &= \rhoCr(g) \begin{bmatrix} \KKCzr \cdots \KKChr \end{bmatrix} D^{\CC_N} V^{-1}\\
    &= \rhoCr(g) \begin{bmatrix} \KKCzr \cdots \KKChr \end{bmatrix} V^{-1} V D^{\CC_N} V^{-1} = \rhoCr(g) \KKCrr {\rhoCr}^{-1}.
\end{align}
\end{subequations}

\maketitle

\section{Verification of Lemma~\ref{lm:condition} on \eqref{eq:irrep-cn-kernel}}
\eqref{eq:irrep-cn-kernel} can be verified to follow Lemma~\ref{lm:condition} as:
\begin{align}
    \label{eq:cn-irrep-proof}
    &\KKCkr(\phi + \theta_{i_1})\\
    &= \diag\big(P(i_1) \KK\big) \beta_k, \quad \text{c.f. \eqref{eq:cn-trivial-proof2}}\\
    &= P(i_1) \diag(\KK) P(i_1)^{-1} \beta_k\\
    &= \rhoCr(g) \KKCkr \psi_{0, k}(g)^{-1}, \quad \text{c.f. \eqref{eq:beta-rotate0}.}
    \label{eq:cn-irrep-lemma1}
\end{align}
We can also verify this for
\begin{equation}
    \oKKCkr = \diag(\oKK) \beta_k.
    \label{eq:irrep-cn-kernel-conj}
\end{equation}

\section{Verification of Lemma~\ref{lm:condition} on \eqref{eq:irrep-dn-kernel}}

First note it is easy to verify that for $i_0 = 0$, i.e. $g = (0, i_1)$, the Lemma~\ref{lm:condition} holds in the same way as \eqref{eq:cn-irrep-proof},
\begin{align}
    \KKDjkr(\phi + \theta) &= \rhoDr(g) \KKDjkr {\psi_{j, k}(g)}^{-1}.
\end{align}
We then generalize \eqref{eq:trivial-exchange} on $\KKCkr$ and $\oKK^{\CC_N}_{k \rightarrow \text{reg}}$ given a reflected action $g = (1, i_1)$:

\begin{subequations}
\begin{align}
    &\KKCkr(-\phi + \theta_{i_1})\\
    &= \diag\big(\KK(-\phi + \theta_{i_1})\big) \beta_k\\
    &= B(i_1) \diag(\oKK) B(i_1)^{-1} \beta_k, \quad \text{c.f. \eqref{eq:beta-rotate1}}\\
    &= B(i_1) \diag(\oKK) \beta_k \psi_{0, k}(g)^{-1}\\
    &= -B(i_1) \diag(\oKKC) \beta_k \psi_{1, k}(g)^{-1}
    \label{eq:irrep-exchange}\\
    %&\oKKCkr(-\phi + \theta_{i_1}) 
    &= B(i_1) \diag(\KKC) \beta_k \psi_{0, k}(g)^{-1}\\
    &= -B(i_1) \diag(\KKC) \beta_k \psi_{1, k}(g)^{-1}.
    \label{eq:irrep-exchange-conj}
\end{align}\end{subequations}

Note that \eqref{eq:irrep-exchange} and \eqref{eq:irrep-exchange-conj} both have two equivalent forms denoted with $\psi_{0, k}(g)$ and $\psi_{1, k}(g)$ respectively. Now we can show $\KK^{\DD_N}_{j, k \rightarrow \text{reg}}$ follows Lemma~\ref{lm:condition} for $j=0$, $i_0=1$, i.e. $g = (1, i_1)$ as:
\begin{subequations}
\begin{align}
    &\KKDjkr(-\phi + \theta_{i_1})\\
    &=\begin{bmatrix}{\KKCkr(-\phi + \theta_{i_1})}^\top & {\oKKCkr}(-\phi + \theta_{i_1})^\top\end{bmatrix}^\top\\
    \begin{split}
        &= \left[\begin{matrix}B(i_1) \diag(\oKKC) \beta_k \psi_{0, k}(g)^{-1} \end{matrix}\right.\\
        &\qquad\qquad \left.\begin{matrix} B(i_1) \diag(\KKC) \beta_k \psi_{0, k}(g)^{-1}\end{matrix}\right]^\top,\\
        &\qquad\qquad\qquad\qquad\qquad\qquad\qquad\text{c.f. \eqref{eq:irrep-exchange}}\\
    \end{split}\\
    \begin{split}
        &= \rhoDr(g) \begin{bmatrix}{\KKCkr}^\top & {\oKKCkr}^\top\end{bmatrix}^\top \psi_{0, k}(g)^{-1},\\
        &\quad\qquad\qquad\qquad\qquad\qquad\qquad\text{c.f. \eqref{eq:irrep-cn-kernel}, \eqref{eq:irrep-cn-kernel-conj}}\\
    \end{split}\\
    &= \rhoDr(g) \KKDjkr \psi_{0, k}(g)^{-1}.
\end{align}
\end{subequations}
The verification is similar for $j=1$, $i_0=1$, i.e. $g = (1, i_1)$:
\begin{subequations}
\begin{align}
    &\KKDjkr(-\phi + \theta_{i_1})\\
    &=\begin{bmatrix}{\KKCkr(-\phi + \theta_{i_1})}^\top & -{\oKKCkr}(-\phi + \theta_{i_1})^\top\end{bmatrix}^\top\\
    \begin{split}
    &= \left[\begin{matrix} -B(i_1) \diag(\oKKC) \beta_k \psi_{1, k}(g)^{-1} \end{matrix}\right.\\
    &\qquad\qquad \left.\begin{matrix} B(i_1) \diag(\KKC) \beta_k \psi_{1, k}(g)^{-1}\end{matrix}\right]^\top, \\
    &\qquad\qquad\qquad\qquad\qquad\qquad\qquad\text{c.f. \eqref{eq:irrep-exchange}}\\
    \end{split}\\
    \begin{split}
    &= \rhoDr(g) \begin{bmatrix}{\KKCkr}^\top & -{\oKKCkr}^\top\end{bmatrix}^\top \psi_{0, k}(g)^{-1},\\
    &\quad\qquad\qquad\qquad\qquad\qquad\qquad\text{c.f. \eqref{eq:irrep-cn-kernel}, \eqref{eq:irrep-cn-kernel-conj}}\\
    \end{split}\\
    &= \rhoDr(g) \KKDjkr \psi_{0, k}(g)^{-1}.
\end{align}
\end{subequations}

\section{Verification of Lemma~\ref{lm:condition} on \eqref{eq:cn-regular-regular-kernel}}

This kernel can be verified as follows for $g = (0, i_1)$:
\begin{subequations}
\begin{align}
    &\KKCrr(\phi + \theta_{i_1})\\
    &=\begin{bmatrix} \KKCzr(\phi + \theta_{i_1}) \cdots \KKChr(\phi + \theta_{i_1}) \end{bmatrix} V^{-1}\\
    \begin{split}
        &= \left[\begin{matrix} \rhoCr(g) \KKCzr \psi_{0, 0}(g)^{-1}, \cdots,\end{matrix}\right.\\
        &\qquad\qquad \left.\begin{matrix}\rhoCr(g) \KKChr \psi_{0, \floor{\frac{N}{2}}}(g)^{-1} \end{matrix}\right] V^{-1},\\
        & \qquad\qquad\qquad\qquad\qquad\qquad\qquad\text{c.f. \eqref{eq:cn-irrep-lemma1}}\\
    \end{split}\\
    &= \rhoCr(g) \begin{bmatrix} \KKCzr \cdots \KKChr \end{bmatrix} D^{\CC_N} V^{-1}\\
    &= \rhoCr(g) \begin{bmatrix} \KKCzr \cdots \KKChr \end{bmatrix} V^{-1} V D^{\CC_N} V^{-1}\\
    &= \rhoCr(g) \KKCrr {\rhoCr}^{-1}.
\end{align}
\end{subequations}

\section{Minimal Implementation}

To illustrate the simplicity of our approach and better explain the tensor operation in the construction of a steerable filter, we list the minimal self-contained PyTorch implementation with only 60 lines of code in this section. Note that this implementation is slightly different from the final version which will be released as open source later.

In the implementation we use $\mathtt{grp} = (0, N)$ and $(1, N)$ to denote $\CC_N$ and $\DD_N$ respectively. $\mathtt{irreps}$, $\mathtt{in\_irreps}$ and $\mathtt{out\_irreps}$ denote irreps by a $M \times 2$ matrix, of which each row denotes an irrep $(j, k)$. The steerable convolution operators $\mathtt{IrrepToRegular}$, $\mathtt{RegularToIrrep}$ and $\mathtt{RegularToRegular}$ preserve the same interface to PyTorch $\mathtt{nn.Conv2d}$. The input and output channels are flattened from a structure of $\mathtt{grp[0] \times \mathtt{grp[1]} \times \mathtt{feature\_multiplicity}}$.

% \begin{listing*}
% \centering
% \inputminted[linenos=true, fontsize=\scriptsize]{python}{minimal.py}
% % \begin{minted}[linenos=true, breaklines, breakafter=d, fontsize=\small]{python} 
% % \input{appendix.py}
% % \end{minted}
% \caption{Minimal implementation of the proposed approach.}
% \label{lst:example}
% \end{listing*}

\begin{lstlisting}[float=*t,language=Python, label={lst:example}, caption={Minimal implementation of the proposed approach.}]
import math; import torch as t; import torch.nn.functional as F; LARGE=1e5

def comp_betas(grp, irreps):                                                    # grp: [int, int], irreps: M*2
    angles = t.arange(grp[1]) * math.pi * 2 / grp[1]                            # grp[1]
    angles = t.ger(angles, irreps[:, 1].float())                                # grp[1]*M
    bases = t.stack([angles.cos(), angles.sin()], 1)[None, :, :, :]             # 1*grp[1]*2*M
    flip = ((-1) ** irreps[:, 0]).float()                                       # M
    flip = flip[None, :] ** t.arange(grp[0])[:, None]                           # grp[0]*M
    return flip[:, None, None, :].mul(bases)                                    # grp[0]*grp[1]*2*M

def comp_affine_grid(grp, ker_size): 
    angles = t.arange(grp[1]) * math.pi * 2 / grp[1]                            # grp[1]
    c, s = angles.cos(), angles.sin()
    rot = t.stack((c, -s, s, c), 1).view(1, grp[1], 2, 2)                       # 1*grp[1]*2*2
    flip = t.Tensor([[[[1, 0], [0, 1]]], [[[1, 0], [0, -1]]]])[:grp[0]]         # grp[1]*1*2*2
    aff = t.cat((t.matmul(flip, rot), t.zeros(grp + (2, 1))), -1)               # grp[0]*grp[1]*2*3
    grid = F.affine_grid(aff.flatten(0, 1), (grp[0] * grp[1], 1) + ker_size, False)
    disk_mask = (grid.norm(dim=-1) > 1).unsqueeze(-1)                           # (grp[0]*grp[1])*H*W
    return grid.masked_fill(disk_mask, LARGE)

class IrrepToRegular(t.nn.Conv2d):
    def __init__(self, grp, in_irreps, out_mult, ker_size, **kwargs):           # grp: [int, int], in_irreps: M*2
        super(IrrepToRegular, self).__init__(len(in_irreps), out_mult, ker_size, **kwargs)
        self.grp = grp
        self.grid = comp_affine_grid(grp, self.kernel_size)                     # (grp[0]*grp[1])*K*K
        self.expand_coeffs = comp_betas(grp, in_irreps)                         # grp[0]*grp[1]*2*len(in_irreps)

    def expand_filters(self, weight):
        C_o, C_i, H, W = weight.shape
        weight = weight.view(1, -1, H, W).expand(self.grp[0] * self.grp[1], -1, -1, -1)
        steered = F.grid_sample(weight, self.grid, 'bilinear', 'zeros', False)  # (grp[0]*grp[1])*(C_o*C_i)*K*K
        steered = steered.view(self.grp + (C_o, 1, -1, H, W))                   # grp[0]*grp[1]*C_o*1*C_i*K*K
        coeffs = self.expand_coeffs.view(self.grp + (1, 2, -1, 1, 1))           # grp[0]*grp[1]*1*2*C_i*1*1
        filters = steered.mul(coeffs)                                           # grp[0]*grp[1]*C_o*2*C_i*K*K
        return filters.flatten(0, 2).flatten(1, 2)                              # (grp[0]*grp[1]*C_o)*(2*C_i)*K*K

    def forward(self, x):
        filters = self.expand_filters(self.weight)
        return F.conv2d(x, filters, self.bias, self.stride,
                self.padding, self.dilation, self.groups)

class RegularToIrrep(IrrepToRegular):
    def __init__(self, grp, in_mult, out_irreps, ker_size, **kwargs):
        super(RegularToIrrep, self).__init__(grp, out_irreps, in_mult, ker_size, **kwargs)

    def expand_filters(self, weight):
        filters = super(RegularToIrrep, self).expand_filters(weight)            # (grp[0]*grp[1]*C_o)*(2*C_i)*K*K
        return filters.permute(1, 0, 2, 3)                                      # (2*C_i)*(grp[0]*grp[1]*C_o)*K*K

class RegularToRegular(IrrepToRegular):
    def __init__(self, grp, in_mult, out_mult, ker_size, **kwargs):
        dct_irreps = t.cartesian_prod(t.arange(grp[0]), t.arange(grp[1] // 2 + 1))
        in_irreps = dct_irreps.repeat_interleave(in_mult, 0)                    # [len(dct_irreps)*in_mult]*2
        super(RegularToRegular, self).__init__(grp, in_irreps, out_mult, ker_size,  **kwargs)
        self.dct_mat = comp_betas(grp, dct_irreps).view(grp[0] * grp[1], -1)[:, :, None]
                                                                                # (grp[0]*grp[1])*(2*len(dct_irreps))*1
    def expand_filters(self, weight):                                           # (grp[0]*grp[1]*out_mult)*
        filters = super(RegularToRegular, self).expand_filters(weight)          #       (2*len(dct_irreps)*in_mult)*K*K
        C_o, C_i, H, W = filters.shape                                          # (grp[0]*grp[1]*out_mult)*
        filters = filters.view(C_o, self.dct_mat.shape[1], -1)                  #       (2*len(dct_irreps))*(in_mult*K*K)
        filters = F.conv1d(filters, self.dct_mat)                               # (grp[0]*grp[1]*out_mult)*
        return filters.view(C_o, -1, H, W)                                      #       (grp[0]*grp[1]*in_mult)*K*K
\end{lstlisting}